\documentclass{article}

\usepackage[final]{nips_2017}

\usepackage[utf8]{inputenc} 
\usepackage[T1]{fontenc}    
\usepackage{hyperref}       
\usepackage{url}            
\usepackage{booktabs}       
\usepackage{amsfonts}       
\usepackage{nicefrac}       
\usepackage{microtype}      
\usepackage{natbib}
\setcitestyle{numbers}
\usepackage{graphicx} 

\usepackage{color}                  
\definecolor{blue}{rgb}{0.0,0.0,1.0}
\definecolor{red}{rgb}{1.0,0.0,0.0}
\definecolor{green}{rgb}{0.0,1.0,0.0}

\title{Flow From Motion: A Deep Learning Approach}

\author{Cem Eteke\thanks{Department of Computer Engineering}
  \qquad
  Hayati Havlucu\thanks{Ar\c{c}elik Research Center for Creative Industries}
  \qquad
  Nisa İrem Kırbaç\footnotemark[2]
  \qquad
  Mehmet Cengiz Onbaşlı\thanks{Department of Electrical and Electronics Engineering} \And
  \qquad
  Aykut Coşkun\footnotemark[2]
  \qquad
  Terry Eskenazi\thanks{Department of Psychology}
  \qquad
  Oğuzhan Özcan\footnotemark[2]
  \qquad
  Barış Akgün\footnotemark[1]\\
  Koç University\\
  Istanbul, Turkey \\
  \texttt{\{ceteke13, hhavlucu16, nkirbac, monbasli,}\\
  \texttt{aykut-coskun, teskenazi, oozcan, baakgun\}@ku.edu.tr} \\
}

\begin{document}

\maketitle

\begin{abstract}
Wearable devices have the potential to enhance sports performance, yet they are not fulfilling this promise. Our previous studies with 6 professional tennis coaches and 20 players indicate that this could be due the lack of psychological or mental state feedback, which the coaches claim to provide. Towards this end, we propose to detect the flow state, mental state of optimal performance, using wearables data to be later used in training. We performed a study with a professional tennis coach and two players. The coach provided labels about the players' flow state while each player had a wearable device on their racket holding wrist. We trained multiple models using the wearables data and the coach labels. Our deep neural network models achieved around 98\% testing accuracy for a variety of conditions. This suggests that the flow state or what coaches recognize as flow, can be detected using wearables data in tennis which is a novel result. The implication for the HCI community is that having access to such information would allow for design of novel hardware and interaction paradigms that would be helpful in professional athlete training.

\end{abstract}

\section{Introduction}
\textbf{Wearable technologies} are rapidly advancing thanks to the developments in consumer electronics, with activity trackers leading the way. However, these devices have yet to fulfill their promise of revolutionizing the way we live. The abandonment rate is relatively high as well. There are many hypotheses out there for why this could be, from perceived ``ugliness'' of the device design to lack of features \cite{Lazar:2015:WWU:2750858.2804288}. Our research agenda is to tackle these both aspects and in this paper, we focus on the latter. 
We are specifically interested in detecting \textbf{players' mental state} by using a wearable device. Our interviews with 6 professional coaches and 20 professional players show that: (1) players do not need/want wearable devices to track their fitness level since they are already self-aware in this respect and (2) their biggest concern is about the tracking and learning to regulate their \textbf{mental states} \cite{havlucu2017understanding}.

In this study, we investigate whether wearable devices, specifically a commercially available activity tracker, can be used to detect more than an activity, such as a psychological state. To reach this goal, we set out to detect the \textbf{flow state}, the mental state of optimal performance of players as they play the game using wearables as a first step towards this end. Csikszentmihalyi \cite{csikszentmihalyi1990flow} defines the flow state as ``\textit{putting oneself in a state of optimal experience, the state in which people are so involved in an activity that nothing else seems to matter}''. The coaches we interviewed claim to be able to observe whether their players are in the flow state or not. This suggests that flow, or what coaches call being \textit{in-the-zone} and \textit{fall}, can be detected, at least for tennis.

Motivated by this, we performed a study involving an experienced coach working with professional tennis players and two of his students. Each player wore a wearable device which recorded data while the coach indicated when the players were in flow or not. Using the coach's labels as targets and the recorded data as inputs, we trained multiple \textbf{machine learning} models. We reached around 98\%
testing accuracy using \textbf{deep neural networks} for a variety of conditions involving multiple data combinations. Our results show that the \textbf{flow state} can be detected using wearables data from an \textbf{Inertial Measuring Unit (IMU)}. To the best of our knowledge, this has never been demonstrated before.

\subsection{Related Work}
Existing work on wearables data in sports mostly concentrate on activity recognition. Um et al. uses \textit{deep learning} to classify exercise motion from large-scale wearable sensor data achieving 92.14\% accuracy with a 3-layered Convolutional Neural Network (CNN) \cite{um2016exercise}. In \cite{chernbumroong2011activity}, 5 activities including sitting, standing, lying, walking, and running are classified using Decision Trees and Artificial Neural Networks using a wrist-worn accelerometer. The authors use 4 separate feature sets from time and frequency domains achieving 94.13\% accuracy with their best models. In \cite{connaghan2011multi}, the authors try to classify tennis strokes - forehand, backhand, and serves - of the players using an IMU which is equipped with accelerometer, gyroscope and magnetometer sensors. They have achieved 90\% accuracy using the fusion of accelerometer, gyroscope and magnetometer sensors. There are other studies that take advantge of IMU sensors in wearable devices. In \cite{spriggs2009temporal}, the authors use a camera and IMU for temporally segmenting human motion into primitive actions. Our work focuses on detecting the flow state as opposed to a specific activity.

Existing literature about flow state detection includes different sensors and are concerned with different tasks.  In \cite{bian2016framework}, the heart rate, interbeat interval, heart rate variability (HRV), high-frequency HRV (HF-HRV), and respiratory rate are argued to be effective indicators of flow. In \cite{de2010psychophysiology}, the authors try to find a relationship between subjective flow and psychophysiological measures while playing piano. They measure arterial pulse pressure, respiration, head movements (via a 3-axis accelerometer) and certain facial muscle activity. They did not find any significant relationship between flow and the head movements. In \cite{nacke2008flow}, the authors use electroencephalography, electrocardiography, electromyography, galvanic skin response, and eye tracking equipment to detect the flow state of participants playing a video game. Our approach of detecting flow in a sports application using IMUs has not been done before.

\begin{table}[t]
	\caption{The collected motion data and their respective ranges.}
    \label{tab:rawrange}
    \centering
    \begin{tabular}{r c c}
   	  & \multicolumn{2}{c}{\small{\textbf{Intervals}}} \\
      \cmidrule(r){2-3}
      & {\small \textbf{Min}}
      & {\small \textbf{Max}} \\
      \toprule
      {\small GravityX}      & -1.0     & 1.0 \\
      {\small GravityY}      & -1.0     & 1.0 \\
      {\small GravityZ}      & -1.0     & 1.0 \\
      \midrule
      {\small AccelerationX} & -17.1031 & 7.0596 \\
      {\small AccelerationY} & -16.2396 & 16.6477 \\
      {\small AccelerationZ} & -16.3296 & 16.8778 \\
      \midrule
      {\small RotationRateX} & -26.3145 & 40.8265 \\
      {\small RotationRateY} & -39.8416 & 32.0937 \\
      {\small RotationRateZ} & -35.2566 & 25.3197 \\
      \midrule
      {\small AttitudeYAW}   & $-\pi$   & $\pi$ \\
      {\small AttitudeROLL}  & $-\pi$   & $\pi$ \\
      {\small AttitudePITCH} & $-\pi/2$ & $\pi/2$ \\
      \bottomrule
    \end{tabular}
\end{table}

\section{Method}
\subsection{Data Collection}


The data was collected using two Apple Watches (Series 2) linked to two Apple iPhones, worn by two tennis players on their racket holding wrists during a match. The flow labels were recorded separately by the players' coach as a binary variable. The duration of the match was 74 minutes. 
The devices start recording before the match begins. In order to capture the data, we use a self-developed application that collects the raw data. The players locate themselves in a corner of the field and raise their hand for 3 seconds. The players then move down the line to the other corner and raise their hand for another 3 seconds. This procedure is done to synchronize the recordings. The motion data was collected for each player for every decisecond (i.e. with 10 Hz) throughout the match. Table \ref{tab:rawrange} summarizes the data and their respective ranges in the recorded data.

The flow labels are recorded on a separate iPhone via another application. The coach observes the match and uses volume up and volume down keys to capture the flow state while speaking out the labels. The flow labels are also written down by one of the researchers next to the coach. This is done to cross-validate the recorded labels from the app in case the coach forgets or mis-presses the buttons on the device. 

As our wearables data, we use motion data captured by an IMU sensor including gravity relative coordinate axes (3D), acceleration along these axes (3D), rotation rate (angular velocity) about these axes (3D), and attitude relative to the magnetic north reference frame (YAW, ROLL, PITCH). The players' heart rates and GPS locations were also recorded to help with flow detection but large chunks of missing data and the poor accuracy of GPS hindered these useless.

\subsection{Data Cleaning and Preprocessing}

\begin{figure}[t]
  \includegraphics[width=1\columnwidth]{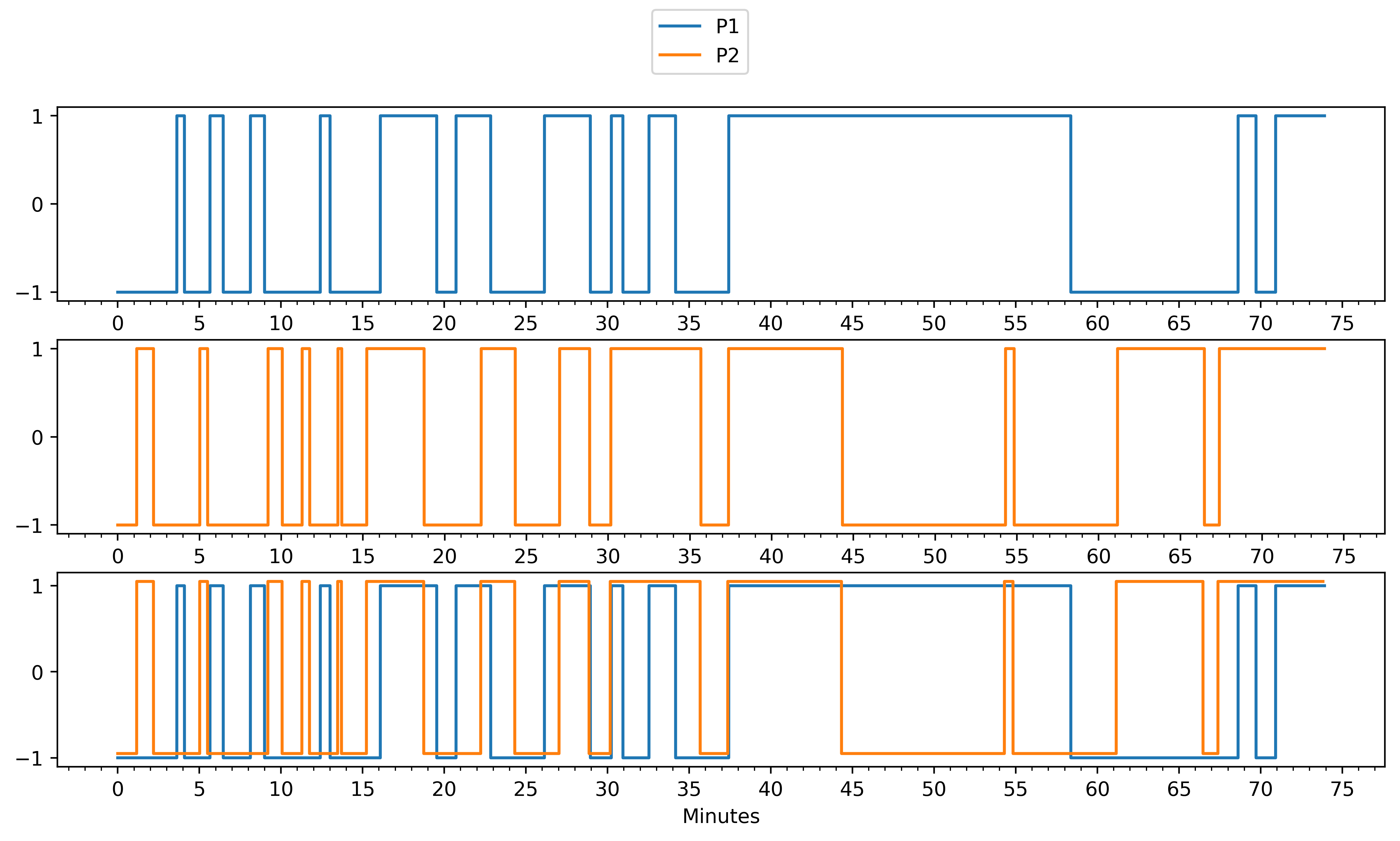}
  \caption{Flow states of the players throughout the match. 1 denotes \textit{flow} and -1 denotes \textit{fall}. In the bottommost plot P2 has been shifted by 0.05 for visualization purposes.(P1: Player 1 - P2: Player 2)}~\label{fig:flows}
\end{figure}

The motion data was pre-processed before being used. The duplicated entries were removed. Next, the data was averaged using a sliding window of size 5. Then, the entries were scaled between -1 and 1. Fig.
The working data for the models contain 44,516 entries for each player, amounting to around 74 minutes. Player 1 and 2 are in flow 51.11\% and 49.95\% of the time respectively. Figure \ref{fig:flows} shows the flow states of the players during the match using the pre-processed data. 1 denotes \textit{flow} and -1 denotes \textit{fall}. 

\section{Learning}

\begin{table}[b]
	\caption{Train/test split combinations used in models. (B: Both Players - P1: Player 1 - P2: Player 2)}	
    \label{tab:datacombination}
    \centering
    \begin{tabular}{r c c}
   	  & \multicolumn{2}{c}{\small{\textbf{Splits}}} \\
      \cmidrule(r){2-3}
      & {\small \textbf{Train}}
      & {\small \textbf{Test}} \\
      \toprule
      {\small B-B}        & 0.9(P1+P2)     & 0.1(P1+P2) \\
      {\small B-P1}       & 0.9P1+1.0P2     & 0.1P1 \\
      {\small B-P2}       & 1.0P1+0.9P2     & 0.1P2 \\
      {\small P1-P1}      & 0.9P1     & 0.1P1 \\
      {\small P1-P2}      & 1.0P1     & 1.0P2 \\
      {\small P2-P1}      & 1.0P2     & 1.0P1 \\
      {\small P2-P2}      & 0.9P2     & 0.1P2 \\
      \bottomrule
    \end{tabular}
\end{table}

The pre-processed data along with coach labels were used to train multiple binary-classifiers. We use 7 different data combinations and corresponding train-test splits to evaluate our models. These combinations are presented in Table~\ref{tab:datacombination}.

\begin{figure}
  \includegraphics[width=1\columnwidth]{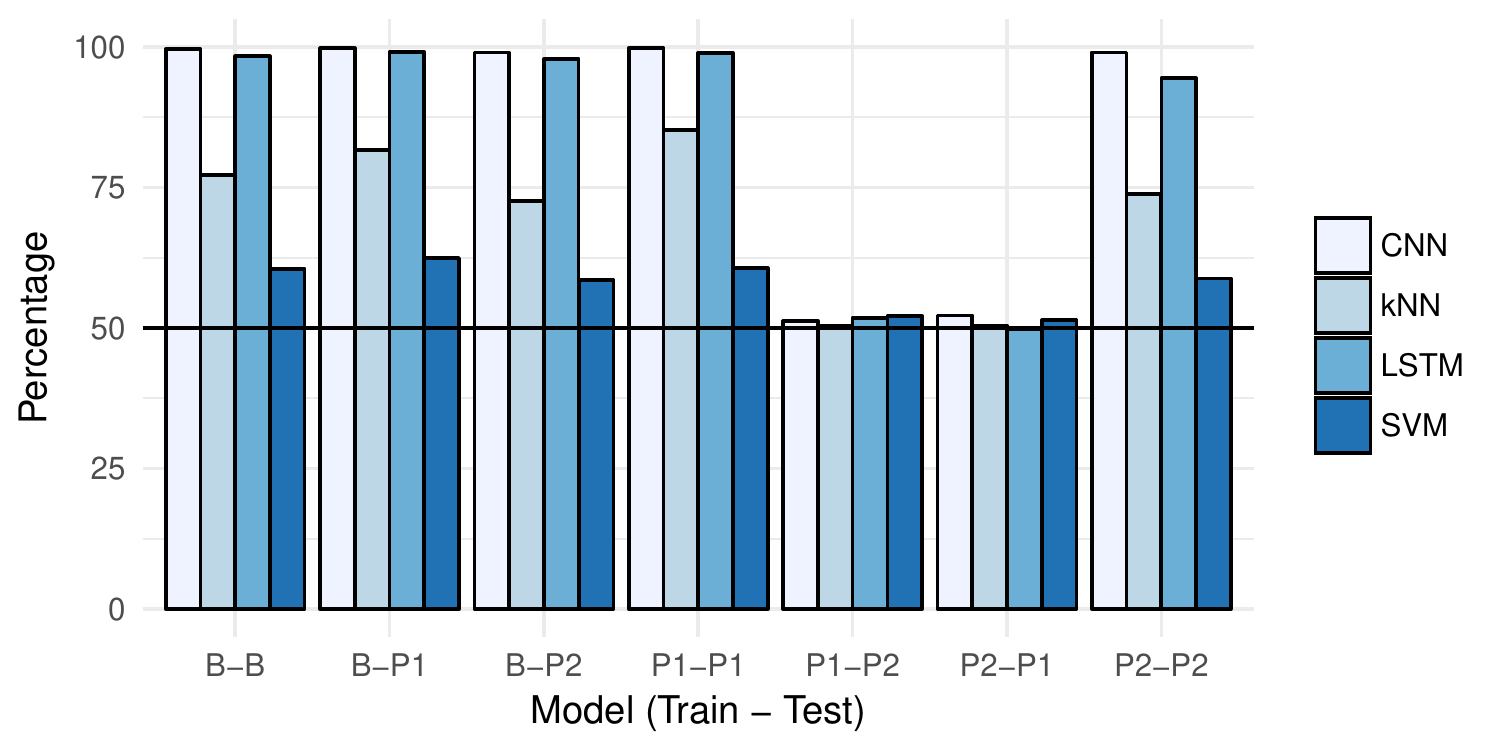}
  \caption{Results of models - the horizontal line represents 50\% accuracy i.e. random chance. (B: Both Players - P1: Player 1 - P2: Player 2)}~\label{fig:modelsum}
\end{figure}

Conventional methods, other than k-Nearest Neighbors (kNN) and random forests performed poorly, barely beating random choice (50\% accuracy) as shown in Figure~\ref{fig:modelsum}. 

Due to the poor performance of the conventional approaches, we used convolutional neural networks (CNNs) and recurrent neural networks (RNN). To further capture the sequential nature of data, our input to these models are formed by combining 10 sequential data points in a sliding windows fashion, resulting in an input dimensionality of $10\times 12$.

The CNN model has three 2D convolutional layers. The first layer has an output size of 128 with a kernel size of $1\times 12$. The second and third layers have size 256 and 512 respectively with kernel size of 1, which results in weight sharing between each time step. 

Activation function of each hidden unit is ReLU and batch normalization is applied after each convolution layer. The last convolutional layer is followed by a fully connected layer of size 128 with ReLU activation function and an output layer of size 2. Softmax function is applied to the output to get flow-state probabilities. 

The RNN model has a single layer of Long Short-Term Memory (LSTM) with hidden unit of size 512 attached to a fully connected layer of size 128 with ReLU activation function and an output layer of size 2. Softmax function is applied to the output to get flow-state probabilities. 

To train both models, we used the Adam optimizer with 0.001 learning rate with no decay and a mini-batch size of 64 to minimize cross-entropy loss.  We included a 0.5 dropout rate before the output layer in both models. We further include a 0.25 dropout rate after the second convolution later in the CNN based model. 

The kNN model uses 1 neighbor and the SVM model uses RBF kernel with $\gamma = 1/12$ and soft-margin cost of $C=1000$ (the latter selected via cross-validation). Figure \ref{fig:modelsum} illustrates the testing accuracy of the models for the data combinations depicted in Table~\ref{tab:datacombination}.


\section{Results}

\textit{Coach's perception of flow can be detected from IMU data in tennis with high accuracy.} The CNN and LSTM models reach around 98\% testing accuracy. Even with a simple approach like kNN, we get around 75\% accuracy. 

\textit{Flow cannot be generalized from a single player.} All the methods are around 50\% when we look at the P1-P2 and P2-P1 parts of the Figure~\ref{fig:modelsum}. This shows that, data from just one player cannot be used to detect flow in others.

\textit{Flow maybe generalized with more player data.} When we look at the B-B, B-P1 and B-P2 parts of the Figure~\ref{fig:modelsum}, we can see that results are as good or better than the P1-P1, and the P2-P2 case. This shows that using more player data may improve flow detection accuracy. This suggests that with more data, we maybe able to detect flow in players we have not trained with but this needs further study.

\textit{Deep neural network based models outperform conventional methods.} The models based on CNN and LSTM have the best results but CNN is slightly better than LSTM for combinations other than P1-P2 and P2-P1. SVM has a very poor performance, barely beating the random chance of 50\% accuracy. The kNN approach with one neighbor is more successful but it is still not competitive\footnote{The story is similar with other conventional methods.}. One reason is that these methods do not account for the sequential nature of the data and utilizing methods such as Hidden Markov Models (HMM) and Conditional Random Fields (CRF) may help. However, our preliminary trials with augmented states (concatenated multiple time steps) and HMMs lacked behind deep models. 

We think that the flow signal is in the IMU data but we need sophisticated models with lots of data to detect it.

\section{Implications and Future Work}
Our end-goal is to be able to detect flow state in professional tennis players. The novel results presented in the previous section strongly support this aim. There are two main directions to take this study; verify flow state detection and advance it and further use the successful detection results to develop devices and interaction paradigms to be used in training to regulate flow state.

Even though the results are highly encouraging, there are still certain challenges to be addressed. These are;
(1) do we need to have training data for a player to be able to detect his/her flow state or can we collect enough data to be able to generalize cross professional tennis players? In other words, can we detect flow in a player we have no training data for? (2) do the movements of professional tennis players change over time and with training such that it affects flow detection in the future? In other words, would the data we collect now be used to detect flow in a player in the future as well?
(3) is what perceived as flow by the coach is really flow and whether this matters or not?  

To address challenges (1) and (2) we need to conduct further studies and collect more data. To address (2) specifically, we need to collect data from the same players over time. Collecting more data is our immediate future work. To address the first half of challenge (3), we need to be able to measure flow directly and to see whether the coach labels are correlated with the measurements. There is no easy way to do this with the current body of flow state knowledge. To address the second half of this challenge, we are planning to follow the first research direction and develop a wearable device for training and see if it works.

A problem that tennis players face is that tennis is a lonely sport and it is hard for them to recover after they lose concentration \cite{havlucu2017understanding}. It is important for these players to train mentally to be able to cope with such difficulties. Not all players get to train with capable coaches or get enough individual training time with them. A wearable device that can help with such mental training would be invaluable. Detecting whether the player is in flow state or not is the first step towards this end. For example, if the device detects that the player goes out-of-flow, it can interact with the user or provide feedback - which is necessary to maintain flow - to help the player get back in flow. We are going to conduct user studies to validate the device and our approach in general.

\section{Conclusion}
In this study, we concentrate on the flow state, mental state of optimal performance, in tennis. We collect flow labels from a professional coach during a tennis match between two of his players and IMU data from the players themselves. We then train several models using this data.

Our findings show that flow, or what the coach perceived as flow, can be detected from IMU data. Most successful methods, two deep learning models, reach around 98\% testing accuracy in a variety of data combinations. The results are the same or better if we have both players' data in the training set. However, one player's data cannot be used to detect flow in the other player. These findings about flow state detection is first in the field.

There are two immediate directions for this study. First, to address data collection and generalization challenges in flow state detection; and second, to develop devices and interaction paradigms to help professional tennis players to train to regulate their flow. We are interested in pursuing both of these directions simultaneously.

\bibliography{references}

\end{document}